\documentclass[10pt,twocolumn,letterpaper]{article}

\usepackage{iccv}
\usepackage{times}
\usepackage{epsfig}
\usepackage{graphicx}
\usepackage{amsmath}
\usepackage{amssymb}

\usepackage{booktabs}
\usepackage{multirow}
\usepackage[normalem]{ulem}
\useunder{\uline}{\ul}{}
\usepackage{subfigure}
\usepackage[T1]{fontenc}
\usepackage{enumitem}

\usepackage{color}
\definecolor{zjw}{rgb}{0.5,0.75,0.6}
\definecolor{new}{rgb}{0.3,0.55,0.4}
\definecolor{jz}{rgb}{0,0,1}

\usepackage[pagebackref=true,breaklinks=true,letterpaper=true,colorlinks,bookmarks=false]{hyperref}
\iccvfinalcopy 

\def\iccvPaperID{4341} 

\ificcvfinal\fi

\begin{document}

\setlength{\abovedisplayskip}{6pt}
\setlength{\belowdisplayskip}{6pt}

\title{Efficient Region-Aware Neural Radiance Fields for High-Fidelity \\ Talking Portrait Synthesis}


\author{
    Jiahe Li$^1$, Jiawei Zhang$^1$, Xiao Bai$^1$\thanks{Corresponding author: Xiao Bai (baixiao@buaa.edu.cn).}, Jun Zhou$^2$, Lin Gu$^{3,4}$ \\
    $^1$School of Computer Science and Engineering, State Key Laboratory of Software Development \\ Environment, Jiangxi Research Institute, Beihang University\\
    $^2$School of Information and Communication Technology, Griffith University\\
    $^3$RIKEN AIP \quad$^4$The University of Tokyo
}


\maketitle
\ificcvfinal\thispagestyle{empty}\fi

\begin{abstract}
This paper presents ER-NeRF, a novel conditional Neural Radiance Fields (NeRF) based architecture for talking portrait synthesis that can concurrently achieve fast convergence, real-time rendering, and state-of-the-art performance with small model size. Our idea is to explicitly exploit the unequal contribution of spatial regions to guide talking portrait modeling. Specifically, to improve the accuracy of dynamic head reconstruction, a compact and expressive NeRF-based Tri-Plane Hash Representation is introduced by pruning empty spatial regions with three planar hash encoders. For speech audio, we propose a Region Attention Module to generate region-aware condition feature via an attention mechanism. Different from existing methods that utilize an MLP-based encoder to learn the cross-modal relation implicitly, the attention mechanism builds an explicit connection between audio features and spatial regions to capture the priors of local motions. Moreover, a direct and fast Adaptive Pose Encoding is introduced to optimize the head-torso separation problem by mapping the complex transformation of the head pose into spatial coordinates. Extensive experiments demonstrate that our method renders better high-fidelity and audio-lips synchronized talking portrait videos, with realistic details and high efficiency compared to previous methods. Code is available at \url{https://github.com/Fictionarry/ER-NeRF}.

\end{abstract}

\section{Introduction}\label{sec:intro}

Audio-driven talking portrait synthesis is an important and challenging issue with several applications such as digital
humans, virtual avatars, film-making, and video conferencing. Over the past few years, many researchers have tackled the task with deep generative models \cite{chen2019hierarchical, prajwal2020wav2lip, thies2020nvp, zhou2020makelttalk, zhou2021pcavs, lu2021lsp, zhang2021facial}. Recently, Neural Radiance Fields (NeRF) \cite{mildenhall2021nerf} is introduced into audio-driven talking portrait synthesis. It provides a new way to learn a direct mapping from the audio feature to the corresponding visual appearance by a deep multi-layer perceptron (MLP). Since then, several studies condition NeRF on audio signals in an end-to-end way \cite{guo2021ad, liu2022semantic, shen2022dfrf, yao2022dfa} or by some intermediate representations \cite{ye2023geneface, Chatziagapi2023lipnerf} to reconstruct a specific talking portrait. Though these vanilla NeRF-based methods have shown great success in the synthesis quality, the inference speed is far from meeting real-time requirements, which seriously limits their practical applications.


Several recent works on efficient neural representation have demonstrated tremendous speedups over vanilla NeRF by replacing part of the MLP network with sparse feature grids \cite{sun2022direct, muller2022instant, chan2022efficient, chen2022tensorf, fang2022fast, cao2023hexplane, fridovich2023k}.
Instant-NGP \cite{muller2022instant} introduces a hash-encoded voxel grid for static scene modeling, allowing fast speed and high-quality rendering with a compact model. 
RAD-NeRF \cite{tang2022rad} first applies this technique to talking portrait synthesis and builds a real-time framework with state-of-the-art performance. 
However, this approach requires a complex MLP-based grid encoder to learn the regional audio-motion mapping implicitly, which limits its convergence and reconstruction quality.

\begin{figure}[t]
\begin{center}
    
\includegraphics[width=1\linewidth]{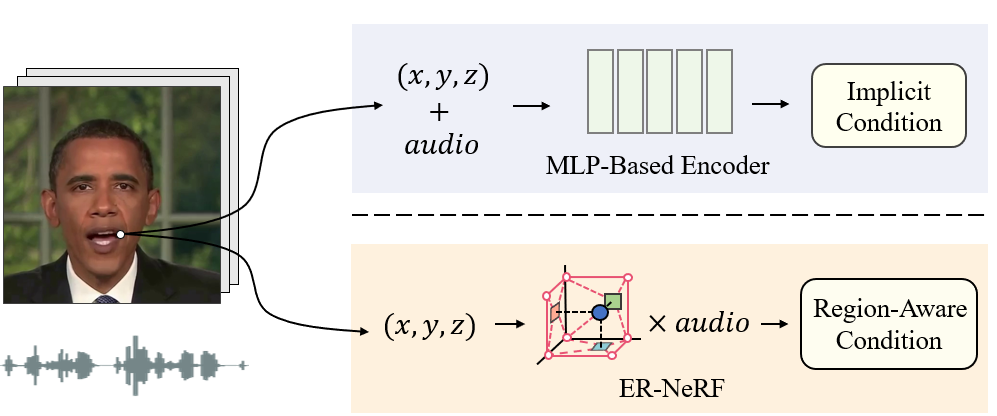}
\end{center}

\caption{Instead of learning the implicit audiovisual relation by an MLP-based encoder, we explicitly attend to the cross-modal interaction between speech audio and spatial regions. Region awareness enables ER-NeRF to render more accurate facial motions.}
\label{fig:mot}
 \vspace{-0.3cm}
\end{figure}

This paper aims to explore a more effective solution for efficient and high-fidelity talking portrait synthesis. 
Based on previous studies, we notice that different spatial regions contribute unequally to representing talking portraits: 
(1) In volumetric rendering, since only the surface regions contribute to representing the dynamic head, most other spatial regions are empty and can be pruned with some efficient NeRF techniques to reduce the training difficulty; 
(2) As the fact that different facial areas have varying associations with speech audio \cite{liu2022semantic}, different spatial regions are inherently related to the audio signal in their own distinct manners and lead to unique audio-driven local motions. Inspired by these observations, we \emph{explicitly exploit the unequal contribution of spatial regions to guide the talking portrait modeling}, and present a novel Efficient Region-aware talking portrait NeRF (\textbf{ER-NeRF}) framework for realistic and efficient talking portrait synthesis, which achieves high-quality rendering, fast convergence, and real-time inference with small model size.

 Our first improvement focuses on the dynamic head representation. Although RAD-NeRF \cite{tang2022rad} has leveraged Instant-NGP to represent the talking portrait and achieves a fast inference, its rendering quality and convergence are hampered by hash collisions when modeling the 3D dynamic talking head. To address this problem, we introduce a \emph{Tri-Plane Hash Representation} that factorizes the 3D space into three orthogonal planes via a NeRF-based tri-plane decomposition \cite{chan2022efficient}. During the factorization, all spatial regions are squeezed onto 2D planes, with the corresponding feature grids pruned. 
 Hence, hash collisions only occur in low-dimensional subspaces and are reduced in number.
 With fewer noises, the network can pay more attention to processing audio features,leading to the capability of reconstructing more accurate head structures and finer dynamic motions.

To capture the regional impact of audio signals, we further explore the relevance between the audio feature and position encoding of the proposed Tri-Plane Hash Representation.
Instead of concatenating the raw features and learning the audiovisual correlation by a large MLP-based encoder, we propose a \emph{Region Attention Module} that adjusts the audio feature to best fit certain spatial regions via a cross-modal attention mechanism. Hence, the dynamic parts of the portrait can acquire more appropriate features to model accurate facial movements, while other static portions remain unaffected by the changing signals. By gaining regional awareness, high-quality and efficient modeling for local motions can be achieved.

Moreover, a simple but effective \emph{Adaptive Pose Encoding} is proposed in our framework to solve the head-torso separation problem. It maps the complex pose transformation onto spatial coordinates and provides a clearer position relation for torso-NeRF to learn its own pose implicitly.

The main contributions of our work are summarized as follows:
\begin{itemize}[leftmargin=*]

\setlength{\topsep}{0pt}
\setlength{\itemsep}{2pt}
\setlength{\parsep}{0pt}
\setlength{\parskip}{2pt}
    \item We introduce an efficient \emph{Tri-Plane Hash Representation} to facilitate dynamic head reconstruction, which also achieves high-quality rendering, real-time inference and fast convergence with a compact model size.
    \item We propose a novel \emph{Region Attention Module} to capture the correlation between the audio condition and spatial regions for accurate facial motion modeling.
    \item Extensive experiments show that the proposed ER-NeRF renders realistic talking portraits with high efficiency and visual quality, which outperforms state-of-the-art methods on both objective evaluation and human studies.
\end{itemize}


\begin{figure*}[!t]
    \centering
    \includegraphics[width=1\linewidth]{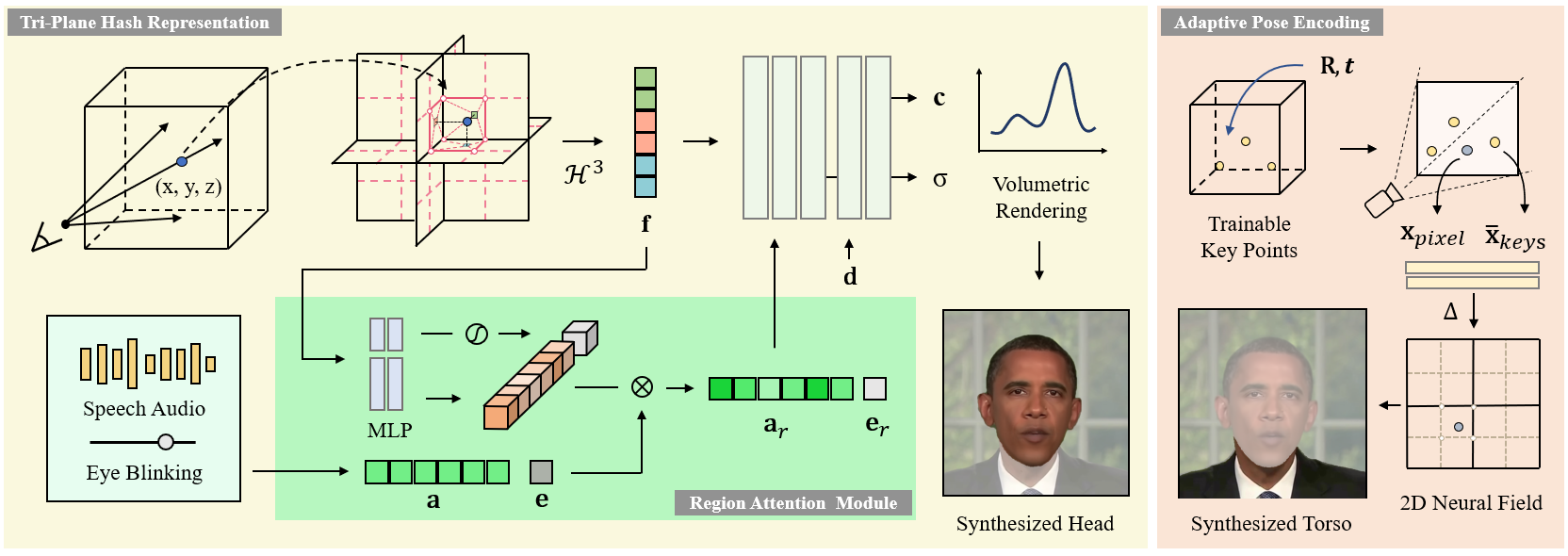}

    \caption{\textbf{Overview of ER-NeRF framework.} The head part of the talking portrait is modeled by the Tri-Plane Hash Representation. A tri-plane hash encoder $\mathcal{H}^3$ is used to encode the 3D coordinate $\mathbf{x}$ into its spatial geometry feature $\mathbf{f}$. The input condition features of speech audio $\mathbf{a}$ and eye blinking $\mathbf{e}$ are reweighted in channel-level with the Region Attention Module and converted to region-aware condition features $\mathbf{a}_r$ and $\mathbf{e}_r$. Then the region-aware features combined with spatial geometry feature $\mathbf{f}$ and the view direction $\mathbf{d}$ are input into an MLP decoder to predict the color $\mathbf{c}$ and density $\sigma$ of the head. The torso part is rendered by another torso-NeRF with the Adaptive Pose Encoding. The corresponding head pose $\mathbf{P}=(\mathbf{R}, \boldmath{t})$ is applied to transform the trainable key points to get their normalized 2D coordinates $\bar{\mathbf{X}}_{keys}$, which conditions a certain 2D Neural Field to predict the torso image.}
    \label{fig:main}
    \vspace{-0.3cm}
\end{figure*}

\vspace{-.2cm}
\section{Related Work}
\vspace{-.1cm}
\noindent\textbf{2D-Based Talking Portrait Synthesis. }
Driving talking portraits by arbitrary speech audio is an active research topic in computer vision and computer graphics. This task aims to reenact the specific person with high image quality and audio-visual consistency. 
Conventional methods~\cite{bregler1997video, brand1999voice} define phoneme-mouth correspondence rules and stitch the mouth shapes. 
Early deep learning-based methods focus on synthesizing the audio-synchronized lip motions for a given facial image \cite{prajwal2020wav2lip, ezzat2002trainable, jamaludin2019you, chen2019hierarchical, wiles2018x2face}. 
Later, to enhance controllability, intermediate representations like facial landmarks and 3D facial models are utilized in several multi-stage methods \cite{thies2020nvp, wang2020mead, zhang2021facial, lu2021lsp}. However, extra errors and information losses would occur in the estimation of these intermediate representations. More recently, diffusion models have been used to improve lip-sync and image quality \cite{yu2022talkingdiff, shen2023difftalk, stypulkowski2023diffused}, but they are slow in inference. Due to the lack of an explicit 3D structure representation, 2D-based methods have drawbacks in the naturalness and consistency of head pose control.

\noindent\textbf{NeRF-based Talking Portrait Synthesis. }
3D vision techniques aim to learn the 3D structure from images and videos relying on multi-view correspondence, and have been widely developed in many areas \cite{wang2021multi, zhang2022learning, wang2022uncertainty, zhang2022revisiting, zhang2020adaptive}.
Recently, Neural Radiance Fields (NeRF)~\cite{mildenhall2021nerf} has been applied to tackle 3D head structure problems in audio-driven talking portrait synthesis. Earlier works~\cite{guo2021ad, yao2022dfa, shen2022dfrf, liu2022semantic} are mainly built on a vanilla NeRF renderer, making them slow and costly for memory. Among them, SSP-NeRF \cite{liu2022semantic} is the first to consider the different impacts of audio on facial areas and adopts a semantic sampling strategy to encourage local motion modeling. 
By applying Instant-NGP \cite{muller2022instant}, RAD-NeRF \cite{tang2022rad} has made huge improvements in visual quality and efficiency. 
 Nevertheless, it requires a complex module to handle audio signals. These end-to-end methods take the whole or part of a large MLP network as the encoder to learn the connection between audio and regions, increasing their complexity and training difficulty. 
Some multi-stage methods \cite{ye2023geneface, Chatziagapi2023lipnerf} pre-train a model to learn the audiovisual relation by intermediate representations, and utilize a NeRF-based renderer for image generation. However, they are inefficient due to the complex architecture.
This paper proposes an efficient NeRF-based method that significantly improves visual quality and audio-lips synchronization.

\noindent\textbf{Efficient Neural Representation. }
 Many reported works focus on the efficiency of NeRF. 
 Recently, several hybrid explicit-implicit representations \cite{chan2022efficient, chen2022tensorf, muller2022instant, sun2022direct, yu2021plenoxels} are proposed for static scene reconstruction and strike a balance between speed and memory cost. In these methods, a high-dimensional scene would be separated and stored into sparse feature grids. Plane-based approaches \cite{ chan2022efficient, chen2022tensorf} factorize the space into multiple low-dimensional planes and vectors to get a compact representation. 
 Instant-NGP \cite{muller2022instant} employs multiple hash tables to store the sparse details in multiresolution, assuming most empty regions have been pruned, which hugely improves memory utilization and rendering quality as well. Although the size of each hash map is usually insufficient for representing all positions in the space, the method does not handle the hash collision explicitly but leaves it to the MLP decoder. These methods are mainly designed for static scenes and are incapable of generating dynamic representation.
 In the field of dynamic NeRFs, current efficient methods are either focused on how to rebuild the scene along the timeline \cite{cao2023hexplane, fridovich2023k, song2022nerfplayer, chen2022tensorf, fang2022fast, wang2022mixed} or can only control some simple deformations \cite{zhang2022controllable}, both of which are unsuitable for modeling audio-driven talking portrait. 
 By leveraging the advantages of the hash and plan-based methods, we introduce an efficient representation for high-quality dynamic head modeling that achieves fast training and inference with small model size.


\section{Method}

\subsection{Preliminaries and Problem Setting} \label{sec: preliminaries}
Given a set of multi-view images and camera poses, NeRF \cite{mildenhall2021nerf} represents a static 3D scene with an implicit function $\mathcal{F}:(\mathbf{x}, \mathbf{d}) \rightarrow (\mathbf{c}, \sigma)$, where $\mathbf{x}=(x, y, z)$ is the 3D spatial coordinate and $\mathbf{d}=(\theta, \phi)$ is the viewing direction. The output $\mathbf{c}=(r, g, b)$ denotes the emitted color and $\sigma$ is the volume density. The color $C(\mathbf{r})$ of one pixel crossed by the ray $\mathbf{r}(t) = \mathbf{o}+t\mathbf{d}$ from camera center $\mathbf{o}$ can be calculated by aggregating the color $\mathbf{c}$ along the ray:
\begin{equation}
    \setlength{\abovedisplayskip}{3pt}
    \setlength{\belowdisplayskip}{3pt}
    \hat{C}(r) = \int _{t_n}^{t_f}\sigma(\mathbf{r}(t)) 
    \cdot 
    \mathbf{c}(\mathbf{r}(t), \mathbf{d}) 
    \cdot
    T(t)dt , 
\end{equation}
where $t_n$ and $t_f$ are the near and far bounds. $T(t)$ is the accumulated transmittance from $t_n$ to $t$:
\begin{equation}
    \setlength{\abovedisplayskip}{3pt}
    \setlength{\belowdisplayskip}{3pt}
    T(t)=\mathrm{exp}(-\int_{t_n}^{t} \sigma(\mathbf{r}(s)) ds).
\end{equation}

In hash grid-based NeRF~\cite{muller2022instant}, a multiresolution hash encoder $\mathcal{H}$ is utilized to encode the spacial point by its coordinate $\mathbf{x}$. Therefore, conditioned with the audio feature $\mathbf{a}$, the basic implicit function of hash NeRF-based audio-driven talking portrait synthesis can be formulated as:
\begin{equation}
    \mathcal{F^A}:(\mathbf{x}, \mathbf{d}, \mathbf{a}; \mathcal{H}) \rightarrow (\mathbf{c}, \sigma).
\end{equation}

In this paper, we adopt the same basic setting as previous NeRF-based works \cite{guo2021ad, liu2022semantic, tang2022rad}. Specifically, we use a few minutes of single-person video as the training data, which is captured from the front view by a static camera. The camera's intrinsic and extrinsic parameters for each frame are calculated from the head poses, which are estimated by a 3DMM model. Audio features are extracted from a pre-trained DeepSpeech \cite{hannun2014deepspeech} model. We also employ an off-the-shelf semantic parsing method to separate the head, torso, and background for various usages. Moreover, we train and render the head and torso separately for acceleration.


\subsection{Tri-Plane Hash Representation} \label{sec: tri-plane hash representation}

 Instant-NGP \cite{muller2022instant} utilizes a set of hash tables to reduce the number of feature grids for efficient neural representation. Inspired by this idea, RAD-NeRF~\cite{tang2022rad} is developed as a real-time and high-quality talking portrait synthesis framework, which leveraged the hash map to represent the small number of surface regions for the portrait head in multiresolution. However, a general 3D hash grid representation is not natively suitable for our task.
 
\begin{figure}[t]
    \subfigure[Static]{
        \includegraphics[width=0.3\linewidth]{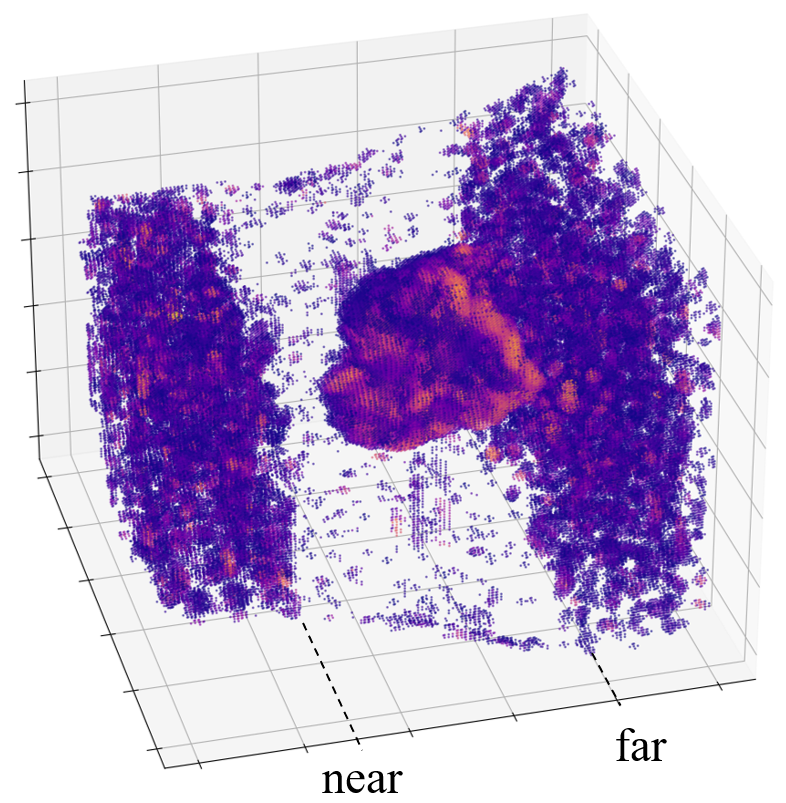}}
    \subfigure[3D hash grid]{
        \includegraphics[width=0.3\linewidth]{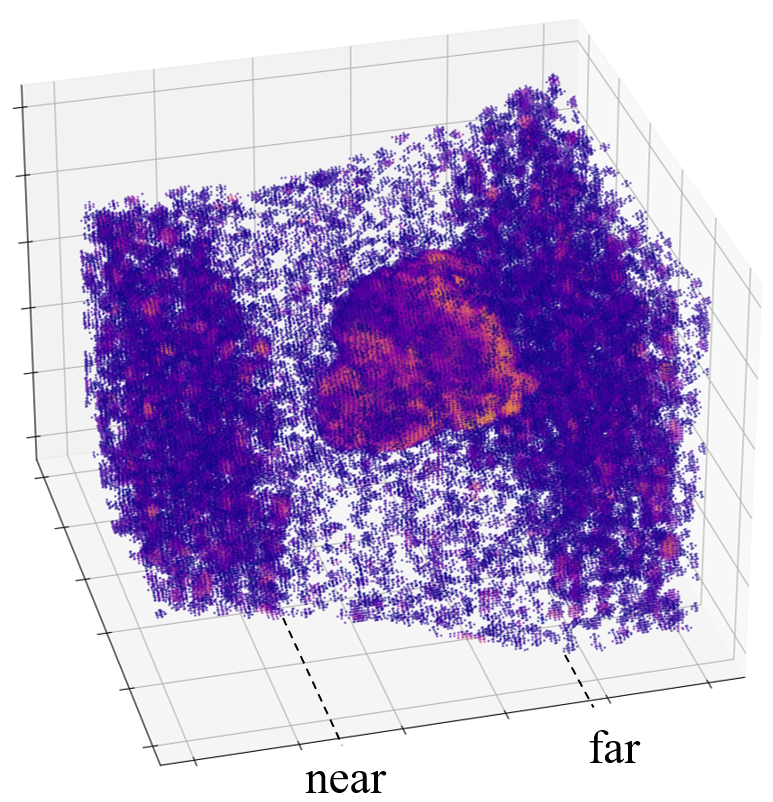}}
    \subfigure[Tri-hash (ours)]{
        \includegraphics[width=0.3\linewidth]{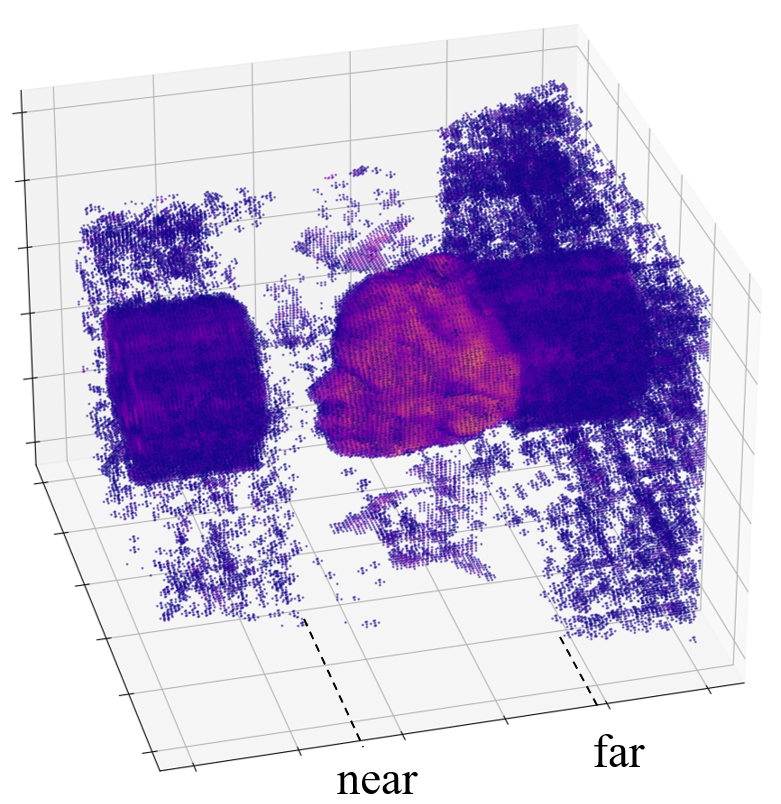}}
   \caption{\textbf{The visualized occupacy grids.} We show the predicted head surfaces according to $\sigma$. \textbf{(a)} 3D hash grid without audio condition. \textbf{(b, c)} 3D hash grid and our tri-plane hash representation conditioned with audio. The MLP decoder of the 3D hash grid becomes overloaded after being required to handle audio features and learn the dynamic motions at the same time, while our representation can still reconstruct the fine surface.}
\label{fig:occ}
\vspace{-0.5cm}
\end{figure}

 A particular problem is the hash collision. Hashing in Instant-NGP treats every position in 3D space equally, which enhances its expressive ability for complex scenes. Nevertheless, the number of hash collisions linearly increases with the number of sampling points, which makes it a burden for the MLP decoder to solve the conflicting gradients. This problem has little effect when reconstructing static scenes, but for talking portrait synthesis, it gets serious when the MLP decoder needs to handle multiple audio features at the same time, as illustrated in Fig.~\ref{fig:occ}.

 \noindent\textbf{Factorization for Hash Grid. } 
Since fewer sampling points always mean lower quality, it's hard to solve this problem by directly reducing the sampling number per ray. Another thinking is to avoid hash collisions from high dimensions. As previous works have proved that a static 3D space of the head can be represented by three 2D tensors \cite{chan2022efficient}, it's possible to squeeze the dynamic talking head into several low-dimensional subspaces with little information loss. From this perspective, we factorize the 3D spatial feature volume into three orthogonal 2D hash grids.

 For a given coordinate $\mathbf{x}=(x, y, z)\in\mathbb{R}^{\mathbf{XYZ}}$, we separately encode its projected coordinates by three 2D-multiresolution hash encoders \cite{muller2022instant}:
 \begin{equation}
    \mathcal{H}^{\mathbf{AB}}: (a, b) \rightarrow \mathbf{f}_{ab}^{\mathbf{AB}}
    \label{eq:planar_hash}
\end{equation}
 where the output $\mathbf{f}_{ab}^{\mathbf{AB}} \in \mathbb{R}^{LF}$ is the plane-level geometry feature for the projected coordinate $(a, b)$ and  $\mathcal{H}^{\mathbf{AB}}$ is the multiresolution hash encoder for plane $\mathbb{R}^{\mathbf{AB}}$, with the number of levels $L$,  feature dimensions per entry $F$. 
 Then we concatenate the results to get the final geometry feature $\mathbf{f}_g \in \mathbb{R}^{3\times LF}$:
\begin{equation}
    \mathbf{f}_{\mathbf{x}} = \mathcal{H}^{\mathbf{XY}}(x,y) \oplus \mathcal{H}^{\mathbf{YZ}}(y,z) \oplus \mathcal{H}^{\mathbf{XZ}}(x,z).
\end{equation}
The symbol $\oplus$ denotes the concatenation operator that concatenates features into a $3\times LF$-channel vector. 

Our proposed factorization significantly reduces hash collision, as now the collision only occurs in 2D planes.  Assuming a common situation that the query rays are almost perpendicular to the frontal plane, the collision can be reduced from $O(R^2N)$ to $O(R^2+2RN)$, where $R^2$ is the number of target pixels and $N$ is the sampling number. With a usual setting of $N=16$ and $R\approx256$ in RAD-NeRF \cite{tang2022rad}, our representation can ideally achieve a $5\times$ reduction in hash collision with the same model size.  This reduction enables the MLP decoder to focus more on processing audio features, leading to improved convergence and dynamic rendering quality. \footref{fn: supp}
 
\noindent\textbf{Overall Head Representation. } The input to the MLP decoder consists of $\mathbf{f}_\mathbf{x}$, the view direction $\mathbf{d}$ and a dynamic condition feature set $\mathcal{D}$ including audio feature. The implicit function of the tri-plane hash representation can be formulated as:
\begin{equation}
    \mathcal{F^H}: (\mathbf{x}, \mathbf{d}, \mathcal{D}; \mathcal{H}^3) \rightarrow (\mathbf{c}, \sigma), 
\end{equation}
where $\mathcal{H}^3: \mathbf{x} \rightarrow \mathbf{f}_{\mathbf{x}}$ denotes a tri-plane hash encoder consisting of all of three planar hash encoders in Eq.~\ref{eq:planar_hash}.

\vspace{-.05cm}
\subsection{Region Attention Module} \label{sec: region attention module}
\vspace{-.05cm}

Dynamic conditions like audio seldom influence the whole portrait equally. Hence, learning how these conditions affect different regions of the portrait is essential for generating natural facial movements. Many previous works \cite{guo2021ad, liu2022semantic, yao2022dfa} ignore this point at the feature level and use some costly approaches to learn the correlation implicitly.  
By leveraging the multi-resolution regional information stored in the hash encoder, we introduce a lightweight region attention mechanism to explicitly fetch the relations between the dynamic feature and different spatial regions.

\noindent\textbf{Region Attention Mechanism. } The region attention mechanism involves an external attention step to calculate the attention vector and a cross-modal channel attention step for reweighting. We aim to connect the dynamic condition feature with the multiresolution geometry feature $\mathbf{f}_\mathbf{x} \in \mathbb{R}^{N}$, which is encoded by the hash encoder $\mathcal{H}$ for a spatial point $\mathbf{x}$. However, since this hierarchical feature is constructed by concatenation, no explicit information flow exists during encoding.

To improve the regional information exchange between different levels of $\mathbf{f}_\mathbf{x}$ efficiently, and further discriminate the importance of audio for each region via the norm of the attention vector, we use a two-layer MLP to capture the global context of the space. Hence it can be explained as the form of external attention mechanism \cite{guo2022externalatt} with two external memory units $M_k$ and $M_v$ for individual levels connection and self-condition query:
\begin{equation}
\setlength{\abovedisplayskip}{3pt}
\setlength{\belowdisplayskip}{3pt}
    \begin{aligned}
        A &= \mathrm{ReLU}(F M_k^T),\\
        V_{out} &= AM_v.
    \end{aligned}
\end{equation}
where vector $\mathbf{f}_\mathbf{x}$ is viewed as an matrix $F\in \mathbb{R}^{N\times1}$.

Then, similar to the channel attention mechanism proposed by Hu et al. \cite{hu2018squeeze}, we treat the resulting feature $V_{out} \in \mathbb{R}^{O\times1}$ as the region attention vector $\mathbf{v} \in \mathbb{R}^O$ to reweight each channel of the dynamic condition feature $\mathbf{q} \in \mathbb{R}^O$. Finally, the output feature vector is:
\begin{equation}
\setlength{\abovedisplayskip}{3pt}
\setlength{\belowdisplayskip}{3pt}
    \mathbf{q}_{out} = \mathbf{v} \odot \mathbf{q}
\end{equation}
where $\odot$ denotes the Hadamard product.
The resulting region-aware feature $\mathbf{q}_{out}$ at each channel is related to hieratical regions where $\mathbf{x}$ is located, since the region attention vector $\mathbf{v}$ includes an informative multi-resolution representation of the space. Therefore, the multi-resolution spatial region can decide which part of the information in $\mathbf{q}$ should be kept or enhanced. 

\noindent\textbf{Speech Audio.} For audio signals, given a query coordinate $\mathbf{x}$ and an audio feature $\mathbf{a} \in \mathbb{R}^A$, we calculate the geometry feature of $\mathbf{x}$ by the tri-plane hash encoder $\mathcal{H}^3$ of our tri-plane hash representation. Then we feed it into a two-layer MLP to generate the region attention vector $\mathbf{v}_{a,\mathbf{x}} \in \mathbb{R}^A$ for audio with the same number of channels $A$. After that, channel-wise attention is applied to $\mathbf{a}$ by $\mathbf{v}_{a,\mathbf{x}}$:
\begin{equation}
\begin{aligned}
    \mathbf{v}_{a,\mathbf{x}} &= \mathrm{MLP}_a(\mathcal{H}^3(\mathbf{x})), \\
    \mathbf{a}_{r, \mathbf{x}} &= \mathbf{v}_{a,\mathbf{x}} \odot \mathbf{a}.
\end{aligned}
\end{equation}
During training, in regions that vary with the audio, the attention vector $\mathbf{v}_{a,\mathbf{x}}$ is optimized for better utilization of the audio feature $\mathbf{a}$. Instead, for the static parts, the audio conditions are considered noises and $\mathbf{v}_{a,\mathbf{x}}$ is going to be a zero vector to help denoising the useless information.

\noindent\textbf{Eye Blinking. } We also apply the mechanism for explicit eye blinking control. We use a scalar to describe the action of eye blinking and regard it as a vector $\mathbf{e}$ with one dimension. Differently, the region attention vector $\mathbf{v}_e \in \mathbb{R}^1$ for eye blinking is output by a sigmoid layer:
\begin{equation}
\begin{aligned}
    \mathbf{v}_{e,\mathbf{x}} &= \mathrm{MLP}_e(\mathcal{H}^3(\mathbf{x})), \\
    \mathbf{e}_{r, \mathbf{x}} &= \mathbf{e} \cdot \mathrm{Sigmoid}(\mathbf{v}_{e,\mathbf{x}}).
\end{aligned}
\end{equation}
The result $\mathbf{e}_{r, \mathbf{x}}$ is scaled by $\mathbf{v}_{e,\mathbf{x}}$ according to its geometry position. In the region of the eyes, $\mathbf{e}_{r, \mathbf{x}}$ conditions the appearance significantly and is close to $\mathbf{e}$ for maximizing its effect. Otherwise, it tends to become $\mathbf{0}$ to reduce the negative interference.

\begin{figure}[t]
\setlength{\abovecaptionskip}{12pt}
\centering
\includegraphics[width=1\linewidth]{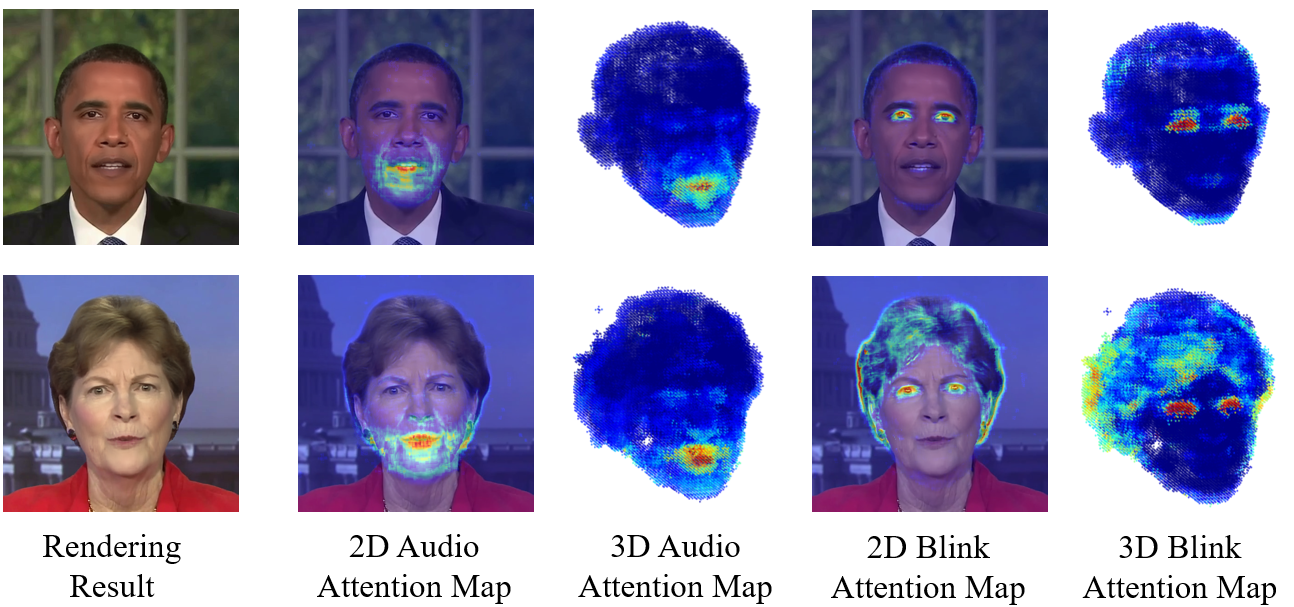}
\caption{\textbf{Visualization of Region Attention Module.} Even if influenced by some uncertain details like fluffy hair, our region attention module successfully captures the relation between dynamic conditions and spatial regions without explicit annotation.}
\label{fig:att}
 \vspace{-0.4cm}
\end{figure}

\vspace{-.1cm}
 \subsection{Training Details} \label{sec: training details}
\vspace{-.05cm}

\noindent\textbf{Adaptive Pose Encoding. }
 To solve the head-torso separation problem, we make an improvement based on previous works \cite{tang2022rad, ye2023geneface}. Instead of directly using the whole image or pose matrix as the condition, we map the complex transformation of the head pose into the coordinates of several key points that have clearer position information, and lead the torso-NeRF to learn an implicit torso pose from these coordinates. 
 
 The encoding process is very simple. We initialize $N$ points in the 3D canonical space with trainable homogeneous coordinates 
  $\mathbf{X}_{keys} \in \mathbb{R}^{4 \times N}$ and apply the head pose $\mathbf{P}= (\mathbf{R}, t)$ to transform the key points $ \hat{\mathbf{X}}_{keys} = \mathbf{P}^{-1}\mathbf{X}_{keys}$.
 Then we project $ \hat{\mathbf{X}}_{keys}$ onto the image plane and get the 2D coordinates $\bar{\mathbf{X}}_{keys}  \in \mathbb{R}^{2 \times N}$ which are the final encoding results to condition the torso-NeRF. 
 In this work, we use $N=3$ and a 2D deformable neural field \cite{tang2022rad} to render the pixel-wise color of the torso .\footnote{Additional descriptions and detailed discussions can be found in the supplementary material. \label{fn: supp}}

 \noindent\textbf{Coarse-to-Fine Optimization.} We apply a two-staged coarse-to-fine training process for better image quality. At the coarse stage, we follow the vanilla NeRF to use the MSE loss for the predicted color $\hat{C}({\mathbf{r}})$ of the image $\mathcal{I}$:
\begin{equation}
\setlength{\abovedisplayskip}{4pt}
\setlength{\belowdisplayskip}{4pt}
    \mathcal{L}_{coarse} = \sum_{\mathrm{i}\in\mathcal{I}}\left \| C({\mathrm{i}}) - \hat{C}({\mathrm{i}}) \right\|^2_2.
\end{equation}
Since MSE loss has a weakness in optimizing sharp details, we then apply an overall finetune with LPIPS loss \cite{zhang2018lpips}. Similar to RAD-NeRF \cite{tang2022rad}, we randomly sample a set of patches $\mathcal{P}$ from the whole image and combine the LPIPS loss by a weight $\lambda$ to enhance details:
\begin{equation}
\setlength{\belowdisplayskip}{3pt}
    \mathcal{L}_{fine} = \sum_{\mathrm{i}\in\mathcal{P}}\left \| C({\mathrm{i}}) - \hat{C}({\mathrm{i}}) \right\|^2_2 + \lambda\ \text{LPIPS}(\hat{\mathcal{P}}, \mathcal{P}).
\end{equation}


\begin{table*}[t]
\resizebox{1\linewidth}{!}{
        \setlength{\tabcolsep}{3.7mm}
        \centering
        \begin{tabular}{lccccccccc}
        \toprule
        Methods & PSNR $\uparrow$ & LPIPS $\downarrow$ & FID $\downarrow$ & LMD $\downarrow$ & AUE $\downarrow$ & Sync $\uparrow$ & Time & FPS & Size (MB) \\
        Ground Truth  & N/A            & 0               & 0              & 0              & 0              & 7.584          & -   & -    & -         \\ \midrule
        Wav2Lip \cite{prajwal2020wav2lip}      & -  & -            & 31.08          & 5.124         & 3.861 & \textbf{8.576} & -   & 19   & $>$400 \\
        PC-AVS \cite{zhou2021pcavs}       & 18.25          & 0.2440          & 101.97         & 4.816          & 3.142          & 8.397          & -   & 32  & $>$500 \\
        AD-NeRF \cite{guo2021ad}      & 30.75          & 0.1034          & 18.60          & 3.345          & 2.201          & 5.205          & 18h & 0.13 & 5.21      \\
        RAD-NeRF \cite{tang2022rad}     & 33.13    & 0.0519          & 12.05          & 2.812          & 2.102          & 5.052          & 5h  & 32   & 11.8      \\ 
        RAD-NeRF\dag    & \textbf{33.26}    & 0.0486          & 12.20          & 2.802          & 1.750          & 5.197          & -  & -   & -      \\ \midrule
        
        ER-NeRF (Ours) & 33.10          & \textbf{0.0291} & \textbf{10.42} & \textbf{2.740} & \textbf{1.629}    & 5.708          & \textbf{2h}  & \textbf{34}   & \textbf{2.51}      \\ \bottomrule 
        \multicolumn{5}{l}{\small\dag \ using AU45 and overall LPIPS finetune. }
        \end{tabular}
    }
    \setlength{\abovecaptionskip}{0cm}
    \caption{\textbf{The quantitative results of the \emph{head reconstruction setting}}. The best results are in \textbf{bold}. 
    Since Wav2Lip can see the ground truth during the self-driven evaluation, we provide another clip of video as the image input. Hence PSNR and LPIPS are not valid. The inference FPS of NeRF-based methods is tested on the Obama dataset \cite{guo2021ad} under the resolution of $450\times450$.}
    \label{tab:setting1}
    \vspace{-0.4cm}
\end{table*}

\vspace{-.2cm}
\section{Experiments}

\vspace{-.1cm}
\subsection{Experimental Settings}\label{sec:setting}
\vspace{-.1cm}

\noindent\textbf{Dataset. } 
  For a fair comparison, the dataset for our experiments is obtained from publicly-released video sets~\cite{guo2021ad, liu2022semantic, shen2022dfrf}. We collect four high-definition speaking video clips with an average length of about 6500 frames in 25 FPS. Each raw video is cropped and resized to $512\times512$ with a center portrait, except the one from AD-NeRF \cite{guo2021ad} with the size of $450\times450$. A pre-trained DeepSpeech model is used to extract the basic audio feature from the speech audio.


\noindent\textbf{Comparison Baselines. }
We compare our method with recent representative one-shot and person-specific models, including Wav2Lip \cite{prajwal2020wav2lip}, PC-AVS \cite{zhou2021pcavs}, NVP \cite{thies2020nvp}, LSP \cite{lu2021lsp} and SynObama \cite{suwajanakorn2017synthesizing}. In addition, we also compare our method with the three end-to-end NeRF-based models: AD-NeRF \cite{guo2021ad}, SSP-NeRF, and RAD-NeRF \cite{liu2022semantic}. Furthermore, we evaluate our method directly on the Ground Truth to provide a clearer comparison.

\noindent\textbf{Implementation Details. }
We implement our method on PyTorch. For a specific portrait, we train the head part for $100,000$ and $25,000$ iterations at the coarse and the fine stage, respectively. In each iteration, we randomly sample a batch of $256^2$ rays from one image. Each 2D hash encoder is set with $L=14, F=1$, and with resolutions from $64$ to $512$. The torso part is trained separately for another $100,000$ iterations. We use AdamW optimizer for both networks with a learning rate of $0.01$ for hash encoders and $0.001$ for other modules. For the control of eye blinking, we choose AU45~\cite{Ekman1978FacialAC} to describe the degree of the action. All experiments are performed on a single RTX 3080Ti GPU. Both the training for the head and torso take about 2 hours.

\vspace{-.1cm}
\subsection{Quantitative Evaluation}
\vspace{-.05cm}

\begin{table}[t]

\resizebox{1\linewidth}{!}{
\centering
\setlength{\tabcolsep}{3mm}
\begin{tabular}{lcccc}
\toprule
             & \multicolumn{2}{c}{Testset A} & \multicolumn{2}{c}{Testset B} \\ \cmidrule(l){2-3}  \cmidrule(l){4-5} 
Methods      & LMD $\downarrow$   & Sync $\uparrow$  & LMD $\downarrow$   & Sync $\uparrow$ \\
Ground Truth & 0            & 6.701          & 0             & 7.309         \\ \midrule
Wav2Lip \cite{prajwal2020wav2lip}      & 6.221        & \textbf{8.378} & \textbf{7.393}& \textbf{8.966}         \\
PC-AVS \cite{zhou2021pcavs}       & 7.112        & 8.087          &  7.722        & 8.565         \\
SynObama \cite{suwajanakorn2017synthesizing}     & 6.540        & 6.802          & -             & -             \\
NVP \cite{thies2020nvp}          & -            & -              & 7.954         & 4.313         \\
LSP \cite{lu2021lsp}          & \textbf{5.905}& 4.287          & 8.122         & 5.843         \\ \midrule
AD-NeRF \cite{guo2021ad}      & \uline{6.192}        & 5.195          & \uline{8.006}        & 4.316         \\
SSP-NeRF \cite{liu2022semantic}     & 6.332        & 5.422          & -             & -             \\
RAD-NeRF \cite{tang2022rad}     & 6.357        & 6.186          & 8.332         & 6.680         \\ 
RAD-NeRF\dag     & 6.339        & 6.119          & 8.355         & 6.392         \\ \midrule
Ours         & 6.254        & \uline{6.242}          & 8.150         & \uline{6.830}         \\ \bottomrule
\multicolumn{5}{l}{\small\dag \ using AU45 and overall LPIPS finetune. }
\end{tabular}
 }
\setlength{\abovecaptionskip}{0cm}
\caption{\textbf{The quantitative results of \emph{lip synthchronization setting}}. The best overall results and the best NeRF-based methods are in \textbf{bold} and {\ul underline}, respectively.}
\label{tab:setting2}
\vspace{-0.5cm}
\end{table}

\begin{figure*}[t]
\centering
   \includegraphics[width=0.92\linewidth]{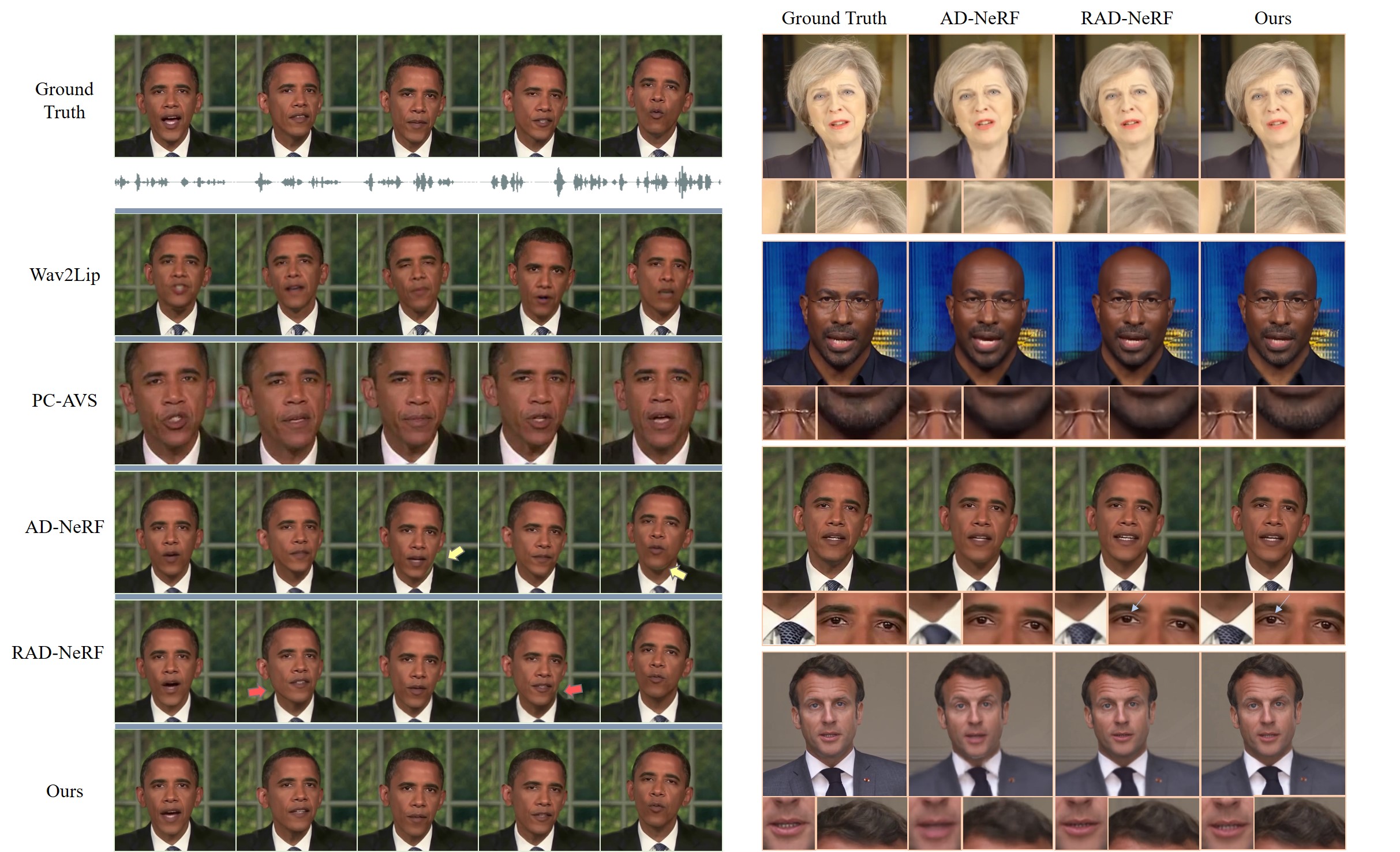}
   \caption{\textbf{The comparison of the key frames and details of generated portraits.} We show the generated results of the baselines \cite{prajwal2020wav2lip, zhou2021pcavs, guo2021ad, tang2022rad} under the head reconstruction setting and the ground truth. For NeRF-based methods, we also synthesize the torso part for evaluation. Please \textbf{zoom in for better visualization.}}
\label{fig:quali}
\vspace{-0.1cm}
\end{figure*}

\begin{table*}[t]
\resizebox{1\linewidth}{!}{
\centering
\begin{tabular}{lcccccccc}
\toprule
Methods           & Wav2Lip \cite{prajwal2020wav2lip} & PC-AVS \cite{zhou2021pcavs} & SynObama \cite{suwajanakorn2017synthesizing} & LSP \cite{lu2021lsp} & NVP \cite{thies2020nvp} & AD-NeRF \cite{guo2021ad}  & RAD-NeRF \cite{tang2022rad} & \textbf{ER-NeRF (Ours)}\quad  \\ \midrule
Lip-sync Accuracy & 2.67     & 2.50     &  3.56   & 2.67   & 2.83   &  3.25    & {\ul 3.81}    & \textbf{4.14}  \\
Image Quality     & 1.92     & 1.83   &  \textbf{4.22}  & 3.83  & 3.75   &  3.33   & 3.69    & {\ul 4.08}  \\
Video Realness    & 1.89      & 1.83      & 3.33  & 2.92  & {\ul 3.50}   & 3.02    & 3.47     & \textbf{3.86}  \\  \bottomrule
\end{tabular}
}
\setlength{\abovecaptionskip}{0cm}
\caption{\textbf{User Study.} The rating is of scale 1-5, the higher the better. We highlight the \textbf{best} and {\ul second best} results.}
\label{tab:user_study}
\vspace{-0.5cm}
\end{table*}

\noindent\textbf{Metrics. }
We employ Peak Signal-to-Noise Ratio (\textbf{PSNR}) to measure the overall image quality and Learned Perceptual Image Patch Similarity (\textbf{LPIPS}) \cite{zhang2018lpips} to measure the details. As we have already used the LPIPS during training, for a fair comparison, an additional feature-based loss Fréchet Inception Distance (\textbf{FID}) \cite{heusel2017fid} is involved for evaluating image quality. We also utilize the landmark distance  (\textbf{LMD}) \cite{chen2018lmd}  and SyncNet confidence score  (\textbf{Sync}) \cite{chung2017syncnet1, chung2017syncnet2}  for lip synchronization and action units error (\textbf{AUE}) \cite{baltrusaitis2018openface, baltruvsaitis2015openface2} to evaluate face motion accuracy.

\noindent\textbf{Comparison Settings. }
In quantitative evaluation, we focus on the synthesized quality of the head. Our comparisons are divided into two settings: 1) The \emph{head reconstruction setting}, where we split each video into training and test dataset to evaluate the reconstruction quality of the head for a specific portrait.  2) The \emph{lip synchronization setting}, where we use the audio track of unseen videos to drive all methods for comparisons in lip synchronization. 

For the first setting, we use all videos in the collected dataset described in Sec.~\ref{sec:setting} and split each video for both training and evaluation. For the second setting, we extract two audio clips from the public demos of NVP and SynObama, named \textbf{Testset A} and \textbf{Testset B}. Due to the lack of pre-trained models and codes for NVP, SynObama, and SSP-NeRF, we also get their generated videos from released demos for evaluation. Following previous works \cite{guo2021ad, liu2022semantic, tang2022rad}, we train our method and other baselines on the Obama dataset released with AD-NeRF \cite{guo2021ad}. For each generated result, we crop and rescaled the facial area into the same size for a fair comparison.

\noindent\textbf{Evaluation Results. }
The results of the \emph{head reconstruction setting} and \emph{lip synchronization setting} are shown in Table \ref{tab:setting1} and Table \ref{tab:setting2}, respectively. It can be observed that: 
(1) In the head reconstruction setting, our method achieves the best reconstruction quality in vision and lip synchronization. Although the one-shot methods (Wav2Lip and PC-AVS) perform best in Sync and can synthesize talking heads without per-scene training, they get poor scores in other metrics, which shows that they cannot accurately reconstruct the specific portrait. For a fair comparison, we also apply the overall LPIPS finetune and AU45 \cite{Ekman1978FacialAC} to RAD-NeRF to enhance its image quality and eye blinking but cause no obvious improvement in image details. Our ER-NeRF performs the best in most metrics while reaching a higher score than other baselines in Sync. The results show that our method can synthesize realistic portraits with high lip-sync accuracy. 
(2) In the lip synchronization setting, our method shows an excellent generalization ability to synthesize lip-sync talking portraits. AD-NeRF and SSP-NeRF encounter an over-smoothing lip movement, leading to a high LMD score but low SyncNet confidence. While getting the highest Sync score among NeRF-based methods, our method exceeds some representative baselines in lip synchronization.
(3) Our method reaches real-time inference, with a faster training time and smaller model size. In Table \ref{tab:setting1}, we report the inference FPS, model size and the time cost for training person-specific models. In comparison, our ER-NeRF achieves the best performance in all three metrics, which demonstrates its high efficiency.

\vspace{-.15cm}
\subsection{Qualitative Evaluation}
\vspace{-.1cm}

\noindent\textbf{Evaluation Results. } 
For an intuitive comparison of the whole portrait, we show the key frames of a clip and details of four portraits in Figure~\ref{fig:quali}. For NeRF-based methods, we synthesize the torso part to evaluate the whole portrait. The result shows that our ER-NeRF renders more details and has the highest personalized lip-sync accuracy. Although Wav2Lip and PC-AVS achieve a high score in Sync, their generated results have an obvious gap from the ground truth. To evaluate the torso part, all three NeRF-based methods render the torso and head separately. AD-NeRF severely suffers from head-torso separation (yellow arrow), and the torso of RAD-NeRF also fails to align with the head sometimes (red arrow). With the same basic representation for the torso as RAD-NeRF, our method demonstrates higher robustness and quality thanks to the design of \emph{Adaptive Pose Encoding}.

\begin{figure}[t]
\setlength{\abovecaptionskip}{3pt}
\centering
   \includegraphics[width=1\linewidth]{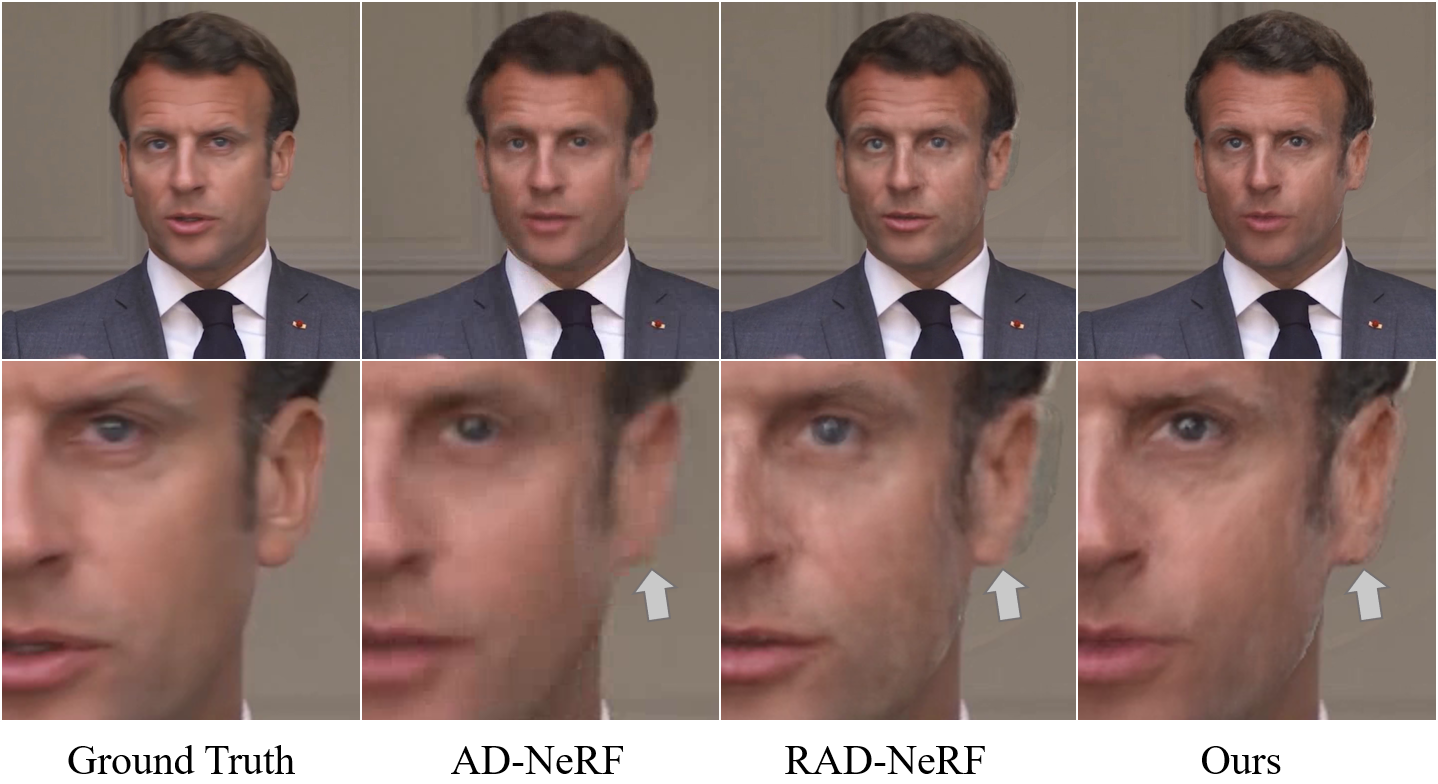}
   \caption{\textbf{Evaluation of the out-of-range pose.} Even with a more compact representation, our method can still render accurate structure at a large rotation angle which is rare in the training video.}
\label{fig:angle}
\vspace{-0.5cm}
\end{figure}

In addition, we also compare results with some out-of-range poses, as shown in Figure \ref{fig:angle}. Despite having pruned most feature grids, our method performs the best in image quality and structure accuracy, which means the robustness and efficiency of our Tri-Plane Hash Representation.

\noindent\textbf{User Study} 
We conducted a user study to better judge the visual quality of the generated heads. We sample 28 generated video clips from the quantitative evaluation, and invite 18 attendees to join the study. The Mean Opinion Scores (MOS) rating protocol is adopted for evaluation and the attendees are required to rate the generated videos from three aspects: (1) Lip-sync Accuracy; (2) Video Realness; (3) Image Quality. The average scores of each method are shown in Table \ref{tab:user_study}. Our ER-NeRF performs the best in lip-sync accuracy and realness. For image quality, only SynObama \cite{suwajanakorn2017synthesizing} gets a higher score than our method, which however relies on a large number of training videos and cannot render in real-time. The results show the excellent visual quality of our method for high-fidelity talking portrait synthesis.

\vspace{-.15cm}
\subsection{Ablation Study}
\vspace{-.12cm}

In this section, we report the ablation study under the head reconstruction setting to prove the effectiveness of our two main contributions. We test settings of different backbones, dynamic feature integration methods and attention targets. The results are shown in Table \ref{tab:ablation} and Table \ref{tab:att_target}.

\noindent\textbf{Representation. } We evaluate three representation backbones on the quality of head reconstruction. The first is an MLP-based network, which is the same as AD-NeRF \cite{guo2021ad}. For grid-based backbones, we compare our tri-hash representation with pure tri-plane in EG3D \cite{chan2022efficient} and the Instant-NGP \cite{muller2022instant} 3D hash grid that is used in RAD-NeRF \cite{tang2022rad}. 
Due to our specialized architecture, the proposed tri-hash representation achieves the best image quality and makes a significant improvement in lip synchronization.

\noindent\textbf{Region Attention Module. } We evaluate the region attention mechanism on two backbones compared with directly concatenating. The results show the enormous impact of our method on modeling accurate motions. Note that by only using our attention mechanism with existing backbones, we can get better scores in both image quality and lip synchronization than current state-of-the-art methods with half of the training time and fewer parameters, which shows the high efficiency of our attention mechanism.

\noindent\textbf{Attention Type. }
 In Table \ref{tab:att_target}, we compare three types of attention for the region attention mechanism: feature-wise and channel-wise. Feature-wise attention scales the entire audio feature with a one-dimensional attention vector, while channel-wise reweights each channel, as described in Section~\ref{sec: region attention module}. The outperforming of channel-wise attention indicates that the proposed region attention mechanism successfully captures the distinct impacts of spatial regions and significantly improves lip motion quality.

\begin{table}[t]
    \resizebox{1\linewidth}{!}{
    \centering
    \begin{tabular}{@{}c|cc|ccccc@{}}
    \toprule
    Backbone                & Concat     & Att.       & PSNR$\uparrow$ & LPIPS$\downarrow$ & LMD$\downarrow$ & AUE$\downarrow$ & Sync$\uparrow$ \\ \midrule
    MLP                     & \checkmark &            & 30.75           & 0.103              & 3.345            & 2.201            & 5.205              \\ \midrule
    \multirow{2}{1.5cm}{\centering Pure \\ Tri-Plane} & \checkmark &            & 32.11           & 0.033        & 2.960            & 1.812   & 4.441              \\
                            &            & \checkmark    & {\ul 33.14}         & {\ul 0.030}        & 2.825   & 1.677      & 5.233     \\ \midrule
    \multirow{2}{*}{iNGP \cite{muller2022instant}}   & \checkmark &            & 33.05           & 0.031              & 2.919            & 1.729            & 4.664              \\
                            &            & \checkmark & 33.12           & {\ul 0.030}              & {\ul 2.810}            & 1.689            & {\ul 5.257}              \\ \midrule
    \multirow{2}{*}{Tri-Hash} & \checkmark &            & \textbf{33.25}  & \textbf{0.029}        & 2.881            & \textbf{1.634}   & 5.123              \\
                            &            & \checkmark & 33.10           & \textbf{0.029}        & \textbf{2.740}   & {\ul 1.646}      & \textbf{5.708}     \\ \bottomrule
    \end{tabular}
    }
    \setlength{\abovecaptionskip}{0cm}
\caption{\textbf{Ablation Study} on Tri-Plane Hash Representation and Region Attention Module.}
\label{tab:ablation}
\vspace{-0.1cm}
\end{table}

\begin{table}[t]
\centering
\resizebox{1\linewidth}{!}{
    \setlength{\tabcolsep}{3mm}
    \begin{tabular}{@{}lccccc@{}}
    \toprule
    Type  & PSNR $\uparrow$ & LPIPS $\downarrow$ & LMD $\downarrow$ & AUE $\downarrow$ & Sync $\uparrow$ \\ \midrule
    Feature-Wise & \textbf{33.14}  & 0.030              & 2.781            & 1.650            & 5.465           \\
    Channel-Wise & 33.10           & \textbf{0.029}     & \textbf{2.740}   & \textbf{1.646}   & \textbf{5.708}  \\ \bottomrule
    \end{tabular}
}
\setlength{\abovecaptionskip}{0cm}
\caption{\textbf{Ablation Study} on types of attention.}
\label{tab:att_target}
\vspace{-0.5cm}
\end{table}

\vspace{-.15cm}
\section{Ethical Consideration} 
\vspace{-.10cm}

We hope our ER-NeRF can enhance interactive experiences and benefits human beings. However, it could be misused for some malicious purposes. As part of our responsibility, we will share our generated results to help develop stronger deepfake detectors. We believe that the responsible use of this technique can promote the healthy growth of both machine learning research and the digital industry.

\vspace{-.15cm}
\section{Conclusion}
\vspace{-.10cm}


In this paper, We propose an efficient and effective framework ER-NeRF for high-quality talking portrait synthesis, mainly consisting of a Tri-Plane Hash Representation and a Region Attention Module. Our framework achieves significant improvement in realistic talking portrait synthesis with higher efficiency. Due to the space limitation, we have put the discussion in the supplementary material. We hope our work can benefit human beings and also inspire more novel conditional NeRF techniques.

\paragraph{Acknowledgments.}
In this work, Jiahe Li, Jiawei Zhang and Xiao Bai are supported by the National Natural Science Foundation of China (No. 62276016), Lin Gu is supported by JST Moonshot R\&D Grant Number JPMJMS2011, Japan.
 
{\small
\bibliographystyle{ieee_fullname}
\bibliography{bib}
}

\clearpage
\appendix

\section{Supplementary Material}
\subsection{Overview}
In the supplemental document, we introduce the details of our torso-nerf with Adaptive Head Encoding, model architecture details, user study details, additional experiments and analysis, ethical considerations, and the discussion of this work. 

\subsection{Torso-NeRF Details \label{sec:torso}}
We combine the proposed Adaptive Pose Encoding and the 2D deformable neural field from RAD-NeRF \cite{tang2022rad} to render the torso part. As described in Section 3.4 of the main paper, we init three points in the 3D canonical space with trainable homogeneous coordinates:
\begin{equation}
    \mathbf{X}_{keys}=(\mathbf{x}_{keys}, \mathbf{y}_{keys}, \mathbf{z}_{keys}, \mathbf{1})^T \in \mathbb{R}^{4 \times 3}.
\end{equation}
For each frame, we form the pose of head $\mathbf{P}$ as:
\begin{equation}
    \mathbf{P} =  \begin{pmatrix}
        \mathbf{R} & \boldsymbol{t} \\
        0 & 1
    \end{pmatrix}
\end{equation}
and apply it to transform the key points:
\begin{equation}
    \hat{\mathbf{X}}_{keys} = \mathbf{P}^{-1}\mathbf{X}_{keys}.
\end{equation}
 where $\hat{\mathbf{X}}_{keys}$ is the transformed coordinates. 
 Then we convert $\hat{\mathbf{X}}_{keys}$ to the ordinary coordinates and project them onto the plane $\mathbf{Z}=1$ to calculate their 2D coordinates $\bar{\mathbf{X}}_{keys} \in \mathbb{R}^{2 \times 3} $ on the imaging plane, where
 \begin{equation}
     \bar{\mathbf{X}}_{keys}(i, j)  = \gamma \cdot \hat{\mathbf{X}}_{keys}(i, j) / \hat{\mathbf{z}}_{keys}(j) ,
 \end{equation}
 and $\gamma$ is the coefficient learned by the network.

\begin{figure*}[t]
    \centering
    \includegraphics[width=1\textwidth]{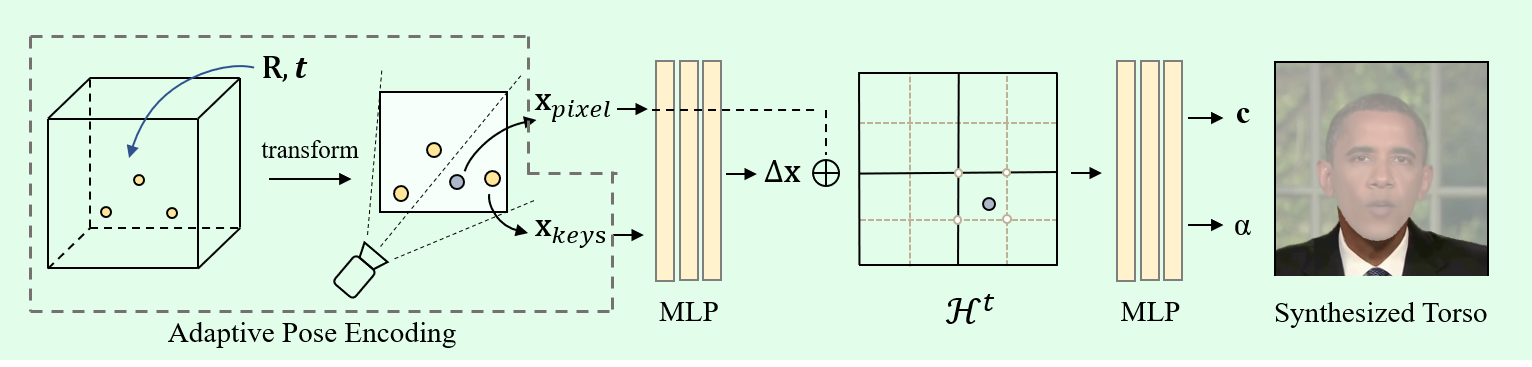}
    \caption{\textbf{Overview of the Torso-NeRF}. }
    \label{fig:torso}
\end{figure*}

 The overview of the torso-NeRF is shown in Figure \ref{fig:torso}. We use $\bar{\mathbf{X}}_{keys}$ to condition the 2D deformable neural field \cite{tang2022rad} for rendering the pixel-wise color and alpha of the torso at the image pixel coordinate $\mathbf{x}_{pixel}$. Specifically, to render the pixel at $\mathbf{x}_{pixel}\in \mathbb{R}^2$ on the image , we firstly feed $\bar{\mathbf{X}}_{keys}$ and the pixel coordinate $\mathbf{x}_{pixel}$ into an MLP, and add the output $\Delta \mathbf{x}$ to $\mathbf{x}_{pixel}$ for a 2D deformation. The deformed coordinate is then encoded by the 2D multiresolution hash encoder $\mathcal{H}^{t}$. Finally, another MLP is used to calculate the pixel-wise transparency $\alpha$ and color $\mathbf{c}_t$. 

 The implicit function of the torso-NeRF can be formulated as:
 \begin{equation}
     \mathcal{F^T}: (\mathbf{x}_{pixel}, \bar{\mathbf{X}}_{keys}; \mathcal{H}^{t}) \rightarrow (\mathbf{c}_t, \alpha)
 \end{equation}

During training, the coordinates $\mathbf{X}_{key}$ can be optimized to gain the ability in representing the implicit relationship between the poses of the head and torso. And due to only linear transformations involved during forwarding, the torso quality is improved without a significant increase in the amount of calculation.

\paragraph{User Study.}
We also conduct a user study to evaluate the synthesized torso part. We invite the attendees to rate the stability and image quality of generated torsos in the \emph{head reconstruction setting}. To compare our method, we selected AD-NeRF \cite{guo2021ad} and RAD-NeRF \cite{tang2022rad} as the baselines since they are the only two NeRF-based methods that can synthesize the torso part and have released their codes. The results are reported in Table \ref{tab:torso_userstudy}. We can observe that our ER-NeRF achieves the best both on Stability and Image Quality by just adding a straightforward encoding step \emph{without any deep neural network}, which demonstrates the high efficiency of our Adaptive Pose Encoding.

\begin{table}[t]
\begin{center}
\begin{tabular}{@{}lccc@{}}
\toprule
Methods       & AD-NeRF & RAD-NeRF & \textbf{ER-NeRF} \\ \midrule
Stability     & 1.33    & 2.89     & \textbf{3.89}             \\
Image Quality & 2.67    & 3.33     & \textbf{4.00}             \\ \bottomrule
\end{tabular} 
\end{center}
\caption{\textbf{User Study of Torso Quality.} The rating is of scale 1-5, the higher the better.}
\label{tab:torso_userstudy}
\end{table}

\subsection{Architecture Details \label{sec:arch}}
\paragraph{Audio Feature Extractor.} In the experiments, we use the pretrained \emph{DeepSpeech} \cite{hannun2014deepspeech} model to extract raw audio features. We then process these features with the same audio attention module as previous NeRF-based works \cite{guo2021ad, shen2022dfrf, tang2022rad}, except for changing the output dimension from 64 to 32.
\vspace{-.3cm}

\paragraph{Region Attention Module.} The speech audio branch utilizes an attention vector MLP with 2 layers and 64 hidden dimensions. Conversely, the eye-blinking branch employs a 2-layer MLP with only 16 hidden dimensions.

\vspace{-.3cm}

\paragraph{Tri-plane Hash Representation.} 
The 2D hash encoders are configured to have 14 resolution levels and a single entry assigned to each level, with a range of multiple resolutions from 64 to 512. The density MLP decoder contains 3 layers, and the color MLP decoder contains 2 layers, both of which have 64 hidden dimensions.

\subsection{User Study Details \label{sec:user}}
The study involves 18 participants with an age range of 20-30 years old. 
To facilitate more accurate judgments, we combine all generated videos and the ground truth into a single high-resolution video. This allows participants to observe all motions simultaneously. To ensure fairness in the comparison process, we assign a number to each generated result instead of identifying them by their method. Participants are asked to evaluate the three perspectives of the generated portraits: (1) Lip-sync Accuracy; (2) Video Realness; (3) Image Quality. To evaluate the torso-NeRF, we additionally invite the attendees to judge two aspects of the synthesized torso: (1) Stability; (2) Image Quality.

\subsection{Tri-Plane Hash Representation \label{sec:tri-hash}}
\paragraph{Complexity of Hash Collision}
 Here we give the proof of the complexity $O(R^2+2RN)$ in Section 3.2 for our Tri-Hash Representation:
 1) For the frontal plane, the projected area is linearly correlated to $R^2$, thus the collision is $O(R^2)$; 
 2) The ideal projected area for the other two side planes is $(\lambda R)R$, where $\lambda$ is an adjustment. But notice only the nearest $N$ points can be sampled at some side areas due to occlusion, so $\lambda R$ is partly correlated to $N$, and the collision is $O(\lambda R^2+RN)$. Overall, $O(R^2+2RN)$ is given.

\begin{table}[t]
\begin{center}
\setlength{\tabcolsep}{4mm}
\resizebox{1\linewidth}{!}{
\begin{tabular}{lccccc}
\toprule
\multirow{2}{*}{\textbf{Grid}} & \multirow{2}{*}{Instant-NGP} & \multicolumn{4}{c}{Tri-Hash}                                                                                     \\ \cmidrule(l){3-6} 
                               &                              & \multicolumn{1}{l}{Frontal} & Side 1                    & Side 2                    & Total                      \\ \midrule
\textbf{Collision}             & 835186                       & 138345                            & \multicolumn{1}{c}{31041} & \multicolumn{1}{c}{26048} & \multicolumn{1}{c}{195434} \\ \bottomrule
\end{tabular}
}
\end{center}
\caption{The number of hash collisions occurring in one feature lookup step on a single grid resolution.}
\end{table}

\paragraph{The Number of Hash Collisions.} 
Here we give the evaluation during one lookup step to directly verify our effect on hash collision reduction. The hashtable size is set to $2^{14}$ and divided by $3$ for each planar grid in our Tri-Hash, with the grid resolution of $512$, the max in the experiment. Adjustments of $1/8$ and $1/4$ are applied due to bilinear interpolation. The point coordinates are scaled up to encourage uniform hashing. In practice, the benefit of our method would be more obvious, since indeed the coordinates cannot be uniformly separated among the hash table and so the overlapping of grids becomes more serious.

\subsection{Additional Experiments \label{sec:add_exp}}
\begin{figure}
    \centering
    \subfigure[Obama]{
        \includegraphics[width=0.48\linewidth]{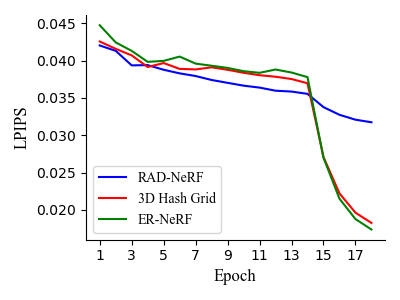}}
    \subfigure[May]{
        \includegraphics[width=0.48\linewidth]{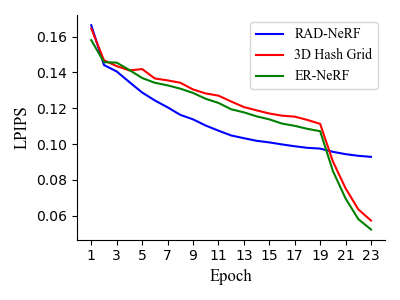}}
    \caption{The validation LPIPS loss on our Obama dataset and May dataset with different architectures. A complex network is much harder to be optimized by the LPIPS finetune and reproduce fine details.}
    \label{fig:lpips}
\end{figure}

\paragraph{LPIPS Finetune.}
It may seem counter-intuitive that the overall LPIPS \cite{zhang2018lpips} finetuning is less effective for RAD-NeRF \cite{tang2022rad} but has a significant impact on the high-frequency details of our ER-NeRF despite having a smaller model size. This phenomenon is likely due to differences in training difficulty. Our ablation study shows that even a simplified architecture with only a 3D hash grid backbone and an audio feature dimension of 32 can reproduce fine details. On the other hand, RAD-NeRF uses a more complex architecture with an additional hash grid and higher-dimensional audio features to improve lip-sync performance, which increases the training difficulty and makes the network harder to optimize. As a result, the LPIPS fine-tuning has a weaker impact on its rendering quality. The variations in LPIPS loss during training are illustrated in Figure \ref{fig:lpips}.

\begin{figure}[t]
    \begin{center}
        
    \includegraphics[width=1\linewidth]{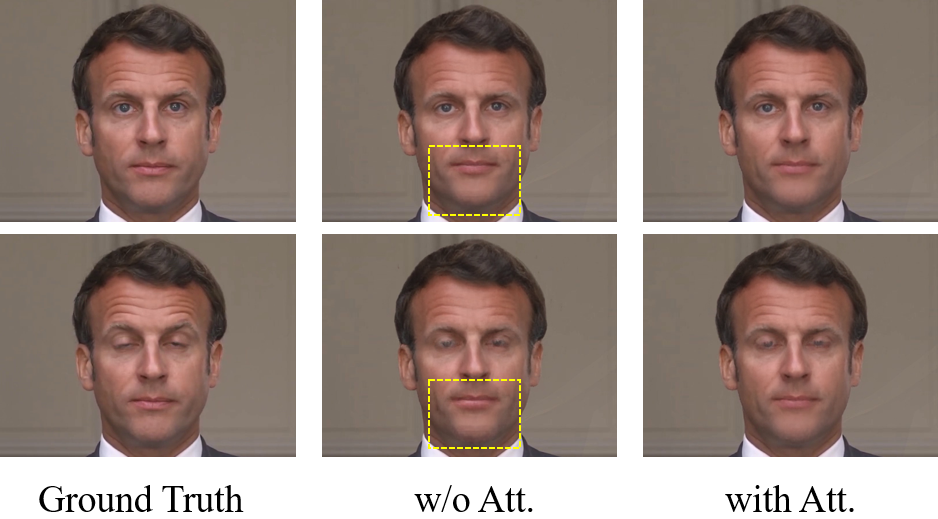}
    \end{center}

    \caption{\textbf{Ablation on Region Attention for Eye Blinking}. Some unnatural facial movements appear when directly concatenating the AU45 with the input to control eye blinking. After applying the proposed region attention mechanism, the robustness has been improved.}
    
    \label{fig:blink}
\end{figure}

\paragraph{Region Attention for Eye Blinking.}
We perform an ablation study on the eye-blinking branch of the Region Attention Module in isolation. When we skip the region attention mechanism and directly concatenate the AU45 with the input of the MLP decoder, some unnatural facial movements appear, like jittering and unreasonable lip movements with eye blinking  (Figure \ref{fig:blink}). This might be due to the module's inability to accurately identify the regional impact of eye blinking and thus learns an incorrect motion mapping with other facial regions. The results indicate that our Region Attention Module can help decouple different semantic motions and improve robustness.

\subsection{Comparison with GeneFace and DFRF} 

In table \ref{tab:setting1_supp} and \ref{tab:setting2_supp}, we have also compared our ER-NeRF with two current SOTA methods GeneFace \cite{ye2023geneface} and DFRF \cite{shen2022dfrf}, both of which are designed for different settings, notably. Meanwhile, since the code of GeneFace is released too close to the submission deadline, it was not taken into the baselines in the main paper. We consider the comparisons not entirely fair for them, and the results are just for reference.


\subsection{Additional Qualitative Comparison}
\begin{figure*}
    \centering
    \includegraphics[width=0.95\linewidth]{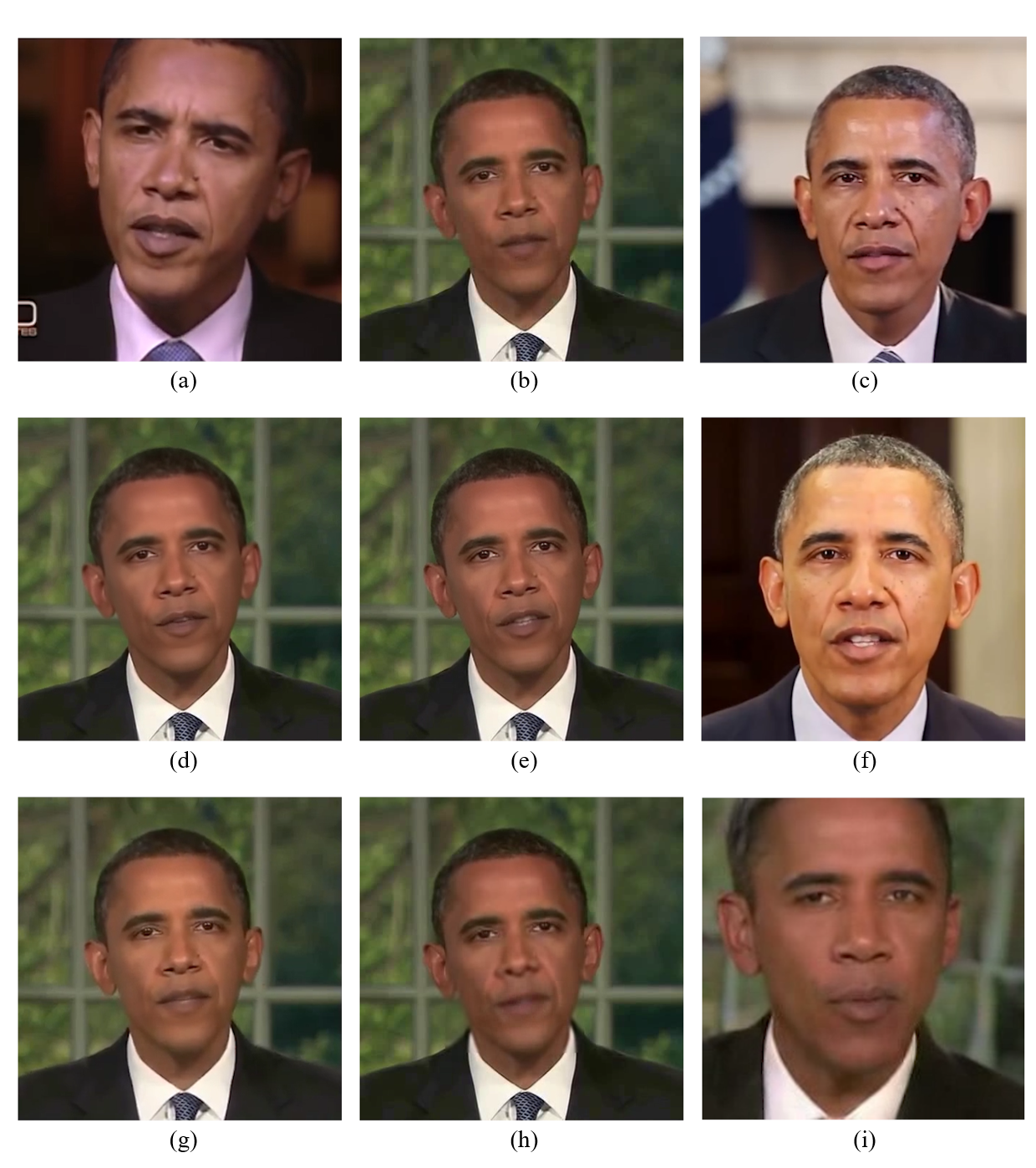}
    \caption{\textbf{Additional Qualitative Comparisons}. We show the synthesized head results of the \emph{lip synchronization setting} on Testset A. \textbf{(a)} Ground truth; \textbf{(b)} AD-NeRF \cite{guo2021ad}; \textbf{(c)} SynObama \cite{suwajanakorn2017synthesizing}; \textbf{(d)} RAD-NeRF \cite{tang2022rad}; \textbf{(e)} \textbf{ER-NeRF (ours)}; \textbf{(f)} LSP \cite{lu2021lsp}; \textbf{(g)} SSP-NeRF \cite{liu2022semantic}; \textbf{(h)} Wav2Lip \cite{prajwal2020wav2lip}; \textbf{(i)} PC-AVS \cite{zhou2021pcavs}.}
    \label{fig:big}
    \vspace{2cm}
\end{figure*}

We show some additional generated key frames on the Testset A under the \emph{lip synchronization setting} with high resolution in Fig. \ref{fig:big}. In this setting, we only synthesize the head part. The results show that our ER-NeRF can outperform most baselines in image quality while retaining a high lip-sync accuracy. We strongly recommend watching our \textcolor[rgb]{0.2, 0.2, 0.8}{supplemental video} for better visualization and more results.

\begin{table}[t]
\begin{center}
\resizebox{1\linewidth}{!}{
    \setlength{\tabcolsep}{2mm}
    \centering
    \begin{tabular}{lccccccccc}
    \toprule
    Methods & PSNR $\uparrow$ & LPIPS $\downarrow$ & FID $\downarrow$ & LMD $\downarrow$ & AUE $\downarrow$ & Sync $\uparrow$ \\ \midrule
    DFRF         & 30.74          & 0.0881           & 13.32          & 3.553         & 2.538          & 4.385         \\
    GeneFace    & 30.24          & 0.0817      & 11.16               & 3.496         & 2.854          & 5.403         \\ \midrule
    
    ER-NeRF (Ours) & \textbf{33.10}          & \textbf{0.0291} & \textbf{10.42} & \textbf{2.740} & \textbf{1.629}    & \textbf{5.708}  \\ \bottomrule 
    \end{tabular}
    }
\end{center}
    \caption{DFRF and GeneFace at the \emph{head reconstruction setting}. }
    \label{tab:setting1_supp}
\end{table}

\begin{table}[t]
\begin{center}
\resizebox{1\linewidth}{!}{
\setlength{\tabcolsep}{3mm}
\begin{tabular}{lcccc}
\toprule
Methods      & A: LMD $\downarrow$   & A: Sync $\uparrow$  & B: LMD $\downarrow$   & B: Sync $\uparrow$ \\ \midrule
DFRF     & 6.551       & 4.854          & 8.126        & 4.127         \\ 
GeneFace     & \textbf{5.465}       & 5.849          & \textbf{7.237}        & 6.275         \\ \midrule
ER-NeRF (Ours)         & 6.254        & \textbf{6.242}          & 8.150         & \textbf{6.830}         \\ \bottomrule
\end{tabular}
 }
\end{center}
\caption{DFRF and GeneFace at the \emph{lip synchronization setting}.}
\label{tab:setting2_supp}
\end{table}


\subsection{Ethics Considerations \label{sec:eth}}
Our proposed ER-NeRF synthesizes high-fidelity talking portraits with accurate lip-audio synchronization. The generated portrait video is highly realistic and difficult for people to distinguish fake from real. We hope it can facilitate a wide range of applications, such as digital humans, video production, and human-computer interaction assistance. On the other hand, however, such techniques may be misused for malicious purposes and make harm. It's significant to tell the users whether a video is real or fake. Recent studies have already achieved success in deepfake detection for face swapping, reenactment and other generating videos \cite{guarnera2020deepfake, zhao2021multi, dolhansky2020deepfake, chen2022self, shiohara2022detecting, dong2022protecting}, but it remains a challenge to discriminate synthesized high-fidelity portraits from recent NeRF-based methods. Besides sharing our generated results to the deepfake detection communication and to help develop more powerful deepfake detectors, we also provide some possible perspectives to fight against the malicious use of talking portrait synthesis:
\begin{itemize}[leftmargin=*]
    \item \textbf{Protect real portrait speech videos}. Since current NeRF-based techniques rely heavily on specific training videos, protection for these real videos is valid to prevent misuse. For example, we can add digital watermarks to the portrait part which can be easily detected even in the generated fake videos.
    
    \item \textbf{Limit the use of deepfake techniques}. Nowadays, little cost of deepfakes leads to an unconstrained use of these techniques. The negative impact of the malicious use of deepfakes can be amplified when they are unintentionally created and shared by the public on social media platforms. Even though the creators may have no malicious intent, the spread of these deepfakes can still have harmful consequences. We suggest the laws should state how to properly make use of these face-generation techniques. On the other hand, the public should also be aware of the potential harm of deepfakes and treat them cautiously.
\end{itemize}


\subsection{Limitation and Future Work \label{sec:limit}}
Compared to the one-shot methods like Wav2Lip \cite{prajwal2020wav2lip}, our method has some advantages in results quality and resolution, however, needs per-scene training when generating new target portraits. Enabling the generative ability may be the target we work for.

Besides, the proposed method has two main limitations. 
 Firstly, our method still encounters a challenge with the small scale of a single training video, leading to a weak lip-audio synchronization with out-of-domain audio, such as some cross-lingual speech or singing voice. Currently, we rely on a pretrained speech recognition model to extract audio features. We have noticed that some recent works \cite{ye2023geneface, Chatziagapi2023lipnerf} employed a pretrained model to enhance their generalizability. In future work, we will consider incorporating priors from large audiovisual datasets to address this limitation.
Secondly, although our method has improved the robustness and image quality of the torso part, there remain some blurry regions. We analyze this may be caused by uncertain movements and the form of representation itself. In future work, we will focus on addressing this issue.

\end{document}